\documentclass{article}

\usepackage{microtype}
\usepackage{graphicx}
\usepackage{subfigure}
\usepackage{enumitem}
\usepackage{amsmath}
\DeclareMathOperator*{\argmax}{argmax}

\setlist{nosep}
\usepackage{booktabs} 

\usepackage{hyperref}

\usepackage[accepted]{icml2023}

\usepackage{amsmath}
\usepackage{amssymb}
\usepackage{mathtools}
\usepackage{amsthm}
\usepackage{multirow}

\usepackage[capitalize,noabbrev]{cleveref}

\theoremstyle{plain}

\theoremstyle{definition}

\theoremstyle{remark}

\usepackage[textsize=tiny]{todonotes}

\icmltitlerunning{Neural Latent Aligner}

\begin{document}

\twocolumn[
\icmltitle{Neural Latent Aligner: Cross-trial Alignment for \\ Learning Representations of Complex, Naturalistic Neural Data}



\icmlsetsymbol{equal}{*}

\begin{icmlauthorlist}
\icmlauthor{Cheol Jun Cho}{ucb}
\icmlauthor{Edward F. Chang}{ucsf}
\icmlauthor{Gopala K. Anumanchipalli}{ucb,ucsf}
\end{icmlauthorlist}

\icmlaffiliation{ucb}{Department of Electrical Engineering and Computer Sciences, University of California, Berkeley, Berkeley, CA, USA}
\icmlaffiliation{ucsf}{Department of Neurological Surgery, University of California, San Francisco, San Francisco, CA, USA}

\icmlcorrespondingauthor{Cheol Jun Cho}{cheoljun@berkeley.edu}

\icmlkeywords{Machine Learning, ICML}

\vskip 0.3in
]



\printAffiliationsAndNotice{\icmlEqualContribution} 

\begin{abstract}

Understanding the neural implementation of complex human behaviors is one of the major goals in neuroscience. To this end, it is crucial to find a true representation of the neural data, which is challenging due to the high complexity of behaviors and the low signal-to-ratio (SNR) of the signals. Here, we propose a novel unsupervised learning framework, Neural Latent Aligner (NLA), to find well-constrained, behaviorally relevant neural representations of complex behaviors. The key idea is to align representations across repeated trials to learn cross-trial consistent information. Furthermore, we propose a novel, fully differentiable time warping model (TWM) to resolve the temporal misalignment of trials. When applied to intracranial electrocorticography (ECoG) of natural speaking, our model learns better representations for decoding behaviors than the baseline models, especially in lower dimensional space. The TWM is empirically validated by measuring behavioral coherence between aligned trials. The proposed framework learns more cross-trial consistent representations than the baselines, and when visualized, the manifold reveals shared neural trajectories across trials.

\end{abstract}

\section{Introduction}
\label{introduction}

Recent advance in neural recording technologies has prompted active development of diverse neuralmodels (e.g., \citealt{pandarinath2018inferring, pei2021neural, kostas2021bendr}), and successful 
applications have provided significant insights into the principles of neural computation and representation \cite{ pandarinath2018inferring, saxena2019towards,vyas2020computation}. Furthermore, these models often provide well-constrained representations that are crucial for developing robust brain-computer interfaces (BCIs) \cite{pandarinath2018latent, dyer2017cryptography, karpowicz2022stabilizing, dabagia2022comparing}. However, application of these models to neural data from naturalistic experiments has been limited, despite the rich implications that such data can offer \cite{kay2008identifying, huth2016natural, silbert2014coupled, chartier2018encoding}. 

A major challenge is that the stimuli and tasks in naturalistic experiments are highly complex, while existing methods have been developed based on conventional experiments with simple task structures. Previous methods assume the low dimensionality of the task structure or simple dynamics \cite{pandarinath2018inferring, ye2021representation}. Neither assumption is applicable to real-world behaviors. However, regardless of complexity, behaviorally relevant representations should encode consistent information when the corresponding behaviors are the same. We leverage this hypothesis to learn representations that exclusively encode information consistent across repeated trials. Another significant obstacle is the temporal misalignment among trials caused by the varying durations of behaviors, which hinders direct comparisons across trials. When properly aligned, a novel pattern can be unveiled \cite{williams2020discovering}. 

\begin{figure*}[ht]
\label{flowchart}
\begin{center}
\centerline{\includegraphics[width=\textwidth]{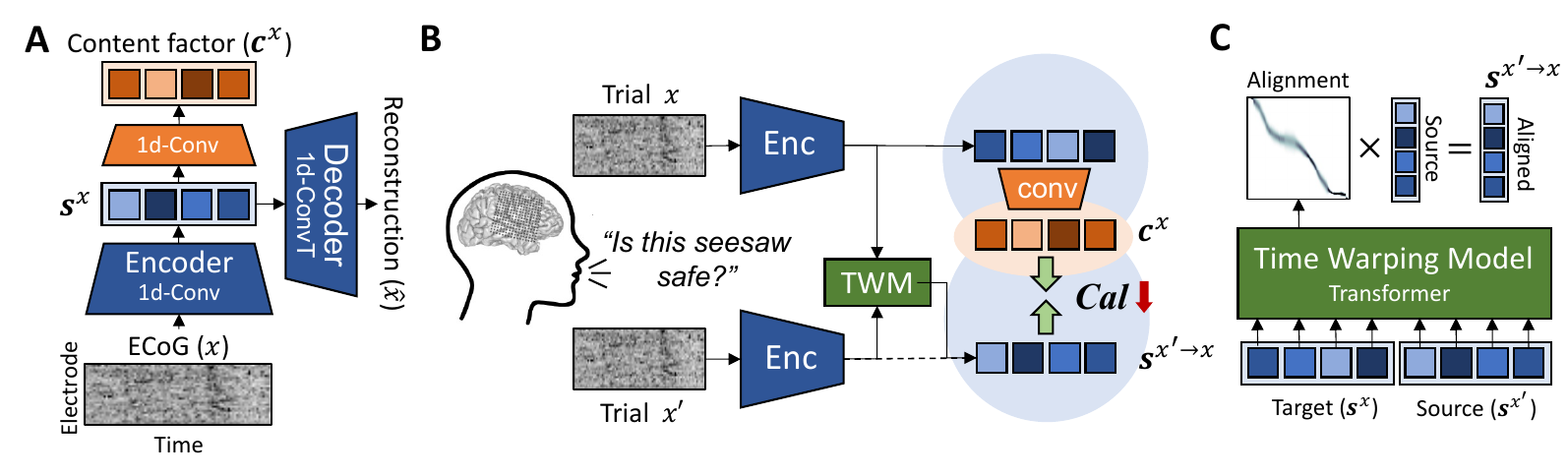}}
\caption{Overview of the proposed framework. \textbf{A}) Sequential autoencoder with 1D CNNs for encoding \(s^x\) and \(c^x\) (content factor). \textbf{B}) Diagram of the proposed cross-trial alignment. Two trials of the same behaviors are provided to the pipeline to minimize the Contrastive alignment loss (\emph{Cal}). \textbf{C}) Time warping model for modeling alignment distribution from source to target.}
\label{framework}
\end{center}
\end{figure*}

Here, we suggest a new unsupervised learning framework, Neural Latent Aligner (NLA), to align latent representations across trials, in both time domain and the representational space. NLA outputs a temporal factor called \emph{content factor}, designed to exclusively represent cross-trial consistent information without noise. To achieve this, we propose a new contrastive loss based on InfoNCE \cite{oord2018representation} by incorporating the repeated trials as an additional sampling dimension. The objective is elaborated in Methods (\S \ref{sec:problem} \& \ref{sec:cal_def}). To address the issue of temporal misalignment, we develop a new time warping model (TWM) with a novel parametrization of temporal alignments. Additionally, we propose two variants of our models, NLA-sDTW and NLA-SUP, which employ distinct approaches for solving the temporal misalignment.

We apply our proposed framework to human-speaking neural data, which consists of high-density intracranial electrocorticography (ECoG) measured from two participants while they read full sentences aloud multiple times. Natural speaking involves the complex coordination of multiple motor tasks, making it suitable to demonstrate the effectiveness of our approach and the limitations of previous methods.

We evaluate our method in three aspects: \textbf{1)}  behavioral relevance of learned representations, \textbf{2)} behavioral coherence of unsupervised alignment, and \textbf{3)} cross-trial consistency. In all three aspects, our model outperforms the baselines: SeqVAE, LFADS \cite{pandarinath2018inferring}, and NDT \cite{ye2021representation} (\S \ref{baselines}). 

Our major contributions are as follows:
\begin{itemize}
\item We propose a novel approach, NLA, for aligning neural data across trials to obtain a true representation from noisy neural data.
\item We introduce a novel parametrization of temporal alignment that can be implemented as a fully differentiable neural network.
\item The content factor of NLA shows the highest correlations with behaviors, in both high- and low-dimensional settings.
\item Our TWM can align trials more coherently than other unsupervised alignment methods. 
\item The content factor exhibits greater consistency across trials compared to baseline representations.
\item The manifold of the content factor reveals shared neural trajectories and dynamical structures across trials.
\end{itemize}

\section{Related Work}
\subsection{Representation learning of neural data}
\label{rw_vae}
Variational autoencoders (VAEs) \cite{kingma2013auto}, and its variants have been widely employed for modeling neural data \cite{pivae,khemakhem2020ivae, pandarinath2018inferring, keshtkaran2022large, liu2021drop}. In particular, sequential VAEs with dynamic priors have been successful in recovering latent dynamical systems of neural processes \cite{pandarinath2018inferring, keshtkaran2022large}. 
Some studies try to learn neural representations invariant to random perturbation (e.g., dropping or time-jittering)  \cite{azabou2021mine, liu2021drop}. Inspired by BERT, predicting randomly masked parts of neural spiking can effectively learn representations of motor neurons \cite{devlin2018bert, ye2021representation}. Contrastive learning has been applied to large-scale EEG data \cite{kostas2021bendr, defossez2022decoding}. However, learning good representations from noisy neural data becomes challenging when the amount of available data is insufficient compared to the task dimensionality.

\subsection{Contrastive learning}
Contrastive learning aims to learn representations that are invariant to variance within positive samples but distinctive from negative samples. This approach has been extensively utilized in the field of self-supervised learning \cite{henaff2020cpc2, chen2020simclr, baevski2020wav2vec, radford2021clip}. The most relevant work to our study is SimCLR \cite{chen2020simclr}, which introduces the concept of generating noisy versions of the data through data augmentation. Here, we obtain the noisy instances from pool of repeated trials rather than from data augmentation.

\subsection{Aligning temporally misaligned representations}


Dynamic Time Warping (DTW) \cite{rabiner1993fundamentals, giorgino2009dtw} is a technique used to find an optimal warping path with minimum frame-wise distance between two sequences. Soft-DTW \cite{cuturi2017soft} is a differentiable version of DTW that allows for backpropagation through the visited paths with weighting, not only the optimal path.

DTW has been applied in representation learning to find features invariant to different durations \cite{haresh2021learning, khaertdinov2022temporal}. Previous works emphasize the importance of incorporating contrast in the loss function to prevent representation degeneration.

Williams and colleagues propose to use TWM to align neural signals and reveal previously unseen patterns \cite{williams2020discovering}. However, their TWM is constrained to be piecewise linear to avoid overfitting caused by high levels of noise, and it requires a specific template for each behavior, limiting its generalizability to unseen behaviors. These limitations make their TWM unsuitable for complex and diverse behaviors.



\section{Methods}
\subsection{Problem formulation}
\label{sec:problem}
Given a behavior (\(y \in \mathcal{Y}\)), a set of neural representations is defined as \(\mathcal{D}_y \), where each data point (\(x \in \mathcal{D}_y \)) is an instance of noisy representations of \(y\) (e.g., raw neural signals). Our objective is to find a mapping (\(\hat{f}\)) that maximizes the mutual information (MI, \(I\)) between the mapped representation (\(f(x)\)) and the true representation of \(y\) (denoted as \( x^*\)) (\ref{eq:problem_definition}).\footnote{Capital letters mean corresponding random variables.}
\begin{equation}
\hat{f} =\argmax\limits_f \: I(f(X); X^*)
\label{eq:problem_definition}
\end{equation}

The above MI maximization can be approximated by minimizing the contrastive loss function (\ref{eq:contrastive_loss1}) \cite{oord2018representation}. 

\begin{equation}
\mathop{\mathbb{E}}_{y\in \mathcal{Y}, x \in \mathcal{D}_y}\Bigr[-\text{log} \: \frac{\text{sim}(f(x),x^*)}{\sum_{\tilde{x} \in \mathcal{D}^*_{\text{neg}}} \text{sim}(f(x), \tilde{x})}\Bigr]
\label{eq:contrastive_loss1}
\end{equation}

This constrastive loss function encourages the representation mapped from noisy signals to be closer to the true signal while being distinctive from negative samples (\(\tilde{x} \in \mathcal{D}^*_{\text{neg}}\)). 

However, since \(x^*\) is not accessible, the loss function is further approximated by additional sampling of \(x'\) (\ref{eq:contrastive_loss2}).
\begin{equation}
\mathop{\mathbb{E}}_{y\in \mathcal{Y}} \Bigr[\mathop{\mathbb{E}}_{(x,x')\in \mathcal{D}_y}  \Bigr[-\text{log}\: \frac{\text{sim}(f(x),x')}{\sum_{\tilde{x} \in \mathcal{D}_{\text{neg}}} \text{sim}(f(x), \tilde{x})}\Bigr]\Bigr]
\label{eq:contrastive_loss2}
\end{equation}

Now, a pair is sampled from \(\mathcal{D}_y\), which requires two dimensions of the sampling pool: 1) a representative set of diverse behaviors (\(\mathcal{Y}\)), and 2) multiple repetitions of each behavior (\(\mathcal{D}_y\)). The datasets described in \S \ref{dataset} satisfy these conditions.

\subsection{Neural Latent Aligner}

\subsubsection{Sequential autoencoder}
\label{sec:archi}
A sequential autoencoder is adopted to model a non-linear projection to latent space. This enables measuring similarities in the latent space of signals, thereby shaping the manifold of learned representations to fulfill the objective. Both the encoder (\(\text{Enc}\)) and decoder (\(\text{Dec}\)) are designed as stacks of three 1D convolutional layers with residual connections (more details can be found in Appendix.\ref{app:A}). The mapping function \(f\) is simply implemented by filtering the outputs of the encoder using two layers of 1D convolution. The autoencoder backbone is trained to minimize the mean squared error of reconstruction (\(\mathcal{L}_{\text{recon}} := \mathop{\mathbb{E}}_{x} \bigr[(x-\text{Dec}(\text{Enc}(x)))^2\bigr]\)).

The content factor, denoted as \(c^x\), is a temporal factor extracted from neural data that captures the representation of our interest (\(c^x:= f(x)\)).
On the other hand, \(s^x\) represents the auto-encoded factor, obtained through the encoding process (\(s^x := \text{Enc}(x)\)). It serves as a surrogate of \(x\) by having all the information from \(x\) including noise. Both factors lie on a \(d\)-dimensional space and form a sequence to represent time series of neural activity (\(c^x_t, s^x_t \in \mathop{\mathbb{R}}^{d}\)). Here, \(d\) is chosen to be sufficiently large (i.e., \(d=256\)) to ensure that \(s^x\) can effectively capture the full
information of the signals. 

\subsubsection{Time warping model}

Considering the alignment of two time series, \(z^A\) and \(z^B\), the probability that the \(j\)-th time-point of \(z^B\) is aligned to the \(i\)-th time point of \(z^A\) is denoted as \(P(i=j|z^A,z^B)\). The probability distribution over \(j\) should be \textbf{1) unimodal} to be one-to-one mapping, and \textbf{2) monotonic} to avoid time-reversing alignment. To address these, we introduce a novel parametrization that satisfies both requirements.

\textbf{1) Unimodality: }The distribution is parametrized as a Gaussian distribution centered at \(\mu_i\) with a variance of \(\sigma_i^2\). 

\textbf{2) Monotonicity: }The means of the distribution are monotonically increasing by estimating non-negative "jumps" between adjacent means (\(\Delta \mu_i := \mu_i - \mu_{i-1}, \Delta \mu_i \geq 0\)). 
\begin{align}
\label{eq:aln_1}
\text{TWM}(z^A, z^B) &= [u^{\Delta\mu}, u^{\sigma}] \\
\label{eq:aln_2}
 \Delta \mu_i &= (L_{B}- \mu_{i-1}) \: \sigma(u^{\Delta\mu}_i) \\
 \label{eq:aln_3}
 \mu_i &= \sum_{k=0}^i \Delta \mu_k \\
 \label{eq:aln_4}
 \sigma_i^2 &= \text{exp}(u^{\sigma}_i)
\end{align}

The alignment variables, \([u^{\Delta\mu}, u^{\sigma}]\), are defined for aligning the source (\(z^B\)), to the target (\(z^A\))(\ref{eq:aln_1}), with the length of \(L_B\) and \(L_A\), respectively. The first variable, \(u^{\Delta\mu}\), is passed through the sigmoid function (\(\sigma(\cdot)\)) and multiplied by the length of the remaining sequence (\( L_B - \mu_{i-1}\)) (\ref{eq:aln_2}). Starting from an initial zero mean (\(\mu_0 = 0\)), the mean at each time point is obtained by accumulating jumps up to that point (\ref{eq:aln_3}). This formulation allows the means to increase monotonically while staying in the range of the source. Lastly, \(u^{\sigma}\) is the logarithm of the variance (\ref{eq:aln_4}).

\begin{align}
\label{eq:aln_6}
P(i=j|z^A,z^B) &\propto \text{exp}\left(\frac{-(j-\mu_i)^2}{2\sigma_i^2}\right) \\
\label{eq:aln_7}
z^{B \rightarrow A}_i &= \sum_{j=1}^{L_B} P(i=j|z^A,z^B) \: z^B_j
\end{align}
The final distribution is parameterized as a Gaussian distribution (\ref{eq:aln_6}), and it is divided by the normalization factor, \(\sum_{k=1}^{L_B}\text{exp}(\frac{-(k-\mu_i)^2}{2\sigma_i^2})\). The final warped sequence is then obtained by the expectation (or weighted sum) (\ref{eq:aln_7}).

\begin{figure}[t]
\begin{center}
\centerline{\includegraphics[width=\columnwidth]{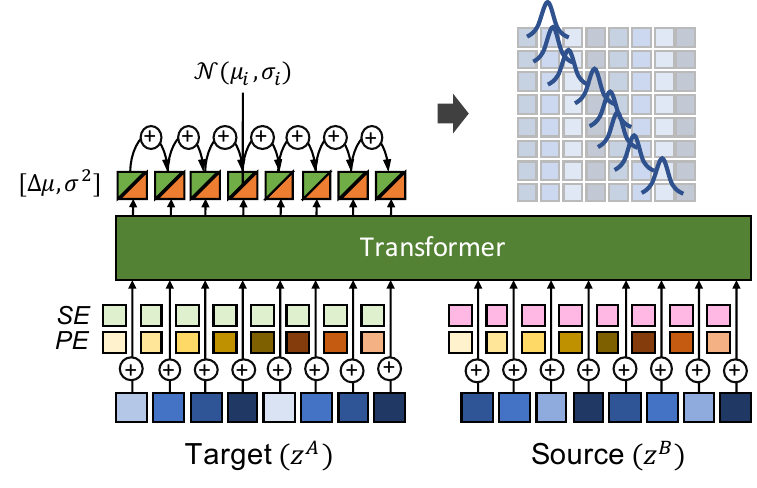}}
\caption{Architecture of the proposed neural time warping model.}
\label{twm}
\end{center}
\end{figure}

This parametrization enables a neural network implementation of \(\text{TWM}(\cdot)\) and for this particular application, we utilize the Transformer  architecture \cite{vaswani2017attention}. The multi-head attention mechanism in Transformer is highly effective in incorporating comparisons within and across sequences. The model takes the target and source sequences as input, and at each time point, two learnable encodings are added: positional encoding to encode the position in the sequence (\(\text{PE}_{i,1\leq i\leq L}\)), and sequence encoding to distinguish between the two sequences (\( \text{SE}_{i,i\in \{1,2\}}\)). An overview of the model is depicted in Figure \ref{twm}.

\subsubsection{Contrastive alignment loss}
\label{sec:cal_def}
As \(x\) and \(x'\) are sequential representations, the similarity is factored by each time point, and summed over the sequence in the log space (\ref{eq:contrastive_loss3}). The exponential of the inner product is used for measuring the similarity.\footnote{The inner product demonstrates more stable training than other metrics. For more details, please refer to Appendix \ref{sec:dist_exp}.} 
\begin{equation}
\mathop{\mathbb{E}}_{y\in \mathcal{Y}} \Bigr[\mathop{\mathbb{E}}_{(x,x')\in \mathcal{D}_y}  \Bigr[-\sum_t^{L_x} \text{log} \:  \frac{\text{sim}(c_t^x,s^{x' \rightarrow x}_t)}{\sum_k^{L_x} \text{sim}(c^x_k,s^{x'\rightarrow x}_t)}\Bigr]\Bigr]
\label{eq:contrastive_loss3}
\end{equation}

The negative samples are also adjusted to be pooled along the sequence of content factors. This modification is crucial for stabilizing the training process and encouraging accurate signal warping.\footnote{Contrasting over warped factors makes training unstable.} We refer to this function as Contrastive alignment loss  \emph{(Cal)}.

\subsubsection{Final loss function}
The final training objective is a weighted sum of the reconstruction loss, the contrastive alignment loss, and additional L2 loss on the content factor (\(\mathcal{L}_{\text{L2}}^{c}\)) (\ref{eq:final_loss}). We set \(\lambda_1 = 0.1, \lambda_2 =0.001\), after some explorations.
\begin{equation}
\mathcal{L}= \mathcal{L}_{\text{recon}} + \lambda_1 \: Cal + \lambda_2 \: \mathcal{L}_{\text{L2}}^{c}
\label{eq:final_loss}
\end{equation}

\subsection{Dataset}

\subsubsection{Dataset description}
\label{dataset}
We apply our framework to ECoG collected by \citet{anumanchipalli2019speech}, where participants read aloud full sentences from MOCHA-TIMIT \cite{mochatimit}. The MOCHA-TIMIT dataset covers a wide range of natural articulations, thus satisfying the first condition of Equation \ref{eq:contrastive_loss2}. Our framework also requires multiple trials of the same behavior to perform the two-fold sampling, sampling a behavior and sampling noisy instances of the behavior. For this purpose, we selected two specific participants, S1 and S2, from the original dataset. S1 spoke 50 sentences for \(9.40\pm0.63\) times, and S2 spoke 450 sentences for \(2.53 \pm 0.50\) times. These sentences are randomly sampled from MOCHA-TIMIT. The duration for each trial is variable to 
 be \(2.29 \pm 0.53\) seconds for S1, and \(2.93 \pm 0.81\) seconds for S2. We split the dataset based on sentences. The total durations for [train, valid, test] sets are [943s, 42s, 93s] for S1, and [2940s, 70s, 327s] for S2.\footnote{The number of sentences in each split [train, valid, test] is [45, 5, 5] for S1 and [400, 10, 40] for S2. For S1, 5 sentences in the valid set are also in the train set but the trials are not overlapped.} No trial of the test sentences is included in the train or valid set. 

The high-density \(16 \times 16\) grids were used to collect the ECoG data from brain regions involved in speech: the ventrals sensorimotor cortex (vSMC), superior temporal gyrus (STG), and inferior frontal gyrus (IFG). The signals are processed to high-gamma amplitudes and downsampled to 200 Hz \cite{chartier2018encoding, anumanchipalli2019speech}. All 256 channels are included and the signals are buffered by 1 second before and after each speaking event.

For the behavioral labels, articulatory kinematic trajectories (AKTs) are extracted by an audio-to-articulatory inversion (AAI) model that is trained on electromagnetic articulography (EMA) corpora \cite{mochatimit, richmond2011mngu}. Each time point is labeled with 12-dimensional articulatory representations, X,Y coordinates of 6 articulators (Figure  \ref{behav_decoding} A). In addition to AKT, we use the frame-wise phoneme labels to evaluate the models. See Appendix \ref{app:data} for detail.


\subsubsection{Online alignment of training dataset}

For each iteration, a batch of sentences is sampled, and then a pair of repeated trials are sampled for each sentence. The pair is not allowed to be identical, since the pool of possible pairs is not large (especially for S2). Then, the 2 seconds of the signals are sampled for each trial of the pair. As this window size does not encompass the entire sentence, we roughly align the trials prior to sampling to ensure the sampled pairs have overlapping content to be aligned.

\label{online}
Assuming we have a roughly estimated alignment between two trials, the onset of the window is randomly sampled from one of the pair, and then, this onset is warped to the other trial to anchor the window of the counterpart.\footnote{Random jitter of \(\pm\) 50 ms is added.} We update the alignment of training data for every 10K of iterations by applying DTW on the content factor. The initial alignment is set as identity function. This online alignment in the middle of training would not be accurate, but it is sufficient to guide the training.
 
\subsection{Baselines}
\subsubsection{Baseline representations}
\label{baselines}
We compare our approach to the following baselines: 
\begin{itemize}
\item \textbf{SeqVAE}: A variational version of our model that minimizes ELBO \cite{kingma2013auto}, instead of the cross-trial alignment. Following previous works \cite{li2018disentangled, zhu2020s3vae, lian2022robust}, we use a learnable dynamic prior modeled by LSTM.

\item \textbf{LFADS}: A sequential VAE based on dynamical system, which is designed to find latent factors of the neural process to infer neural population dynamics \cite{pandarinath2018inferring, keshtkaran2022large}. 
\item \textbf{NDT}: Neural Data Transformer (NDT) is a masked autoencoder for neural data \cite{ye2021representation}. The model is composed of Transformers that learns representations by predicting randomly masked inputs.
\end{itemize}

The rationale for this selection is that \textbf{1)} SeqVAE shares the backbone autoencoder with NLA, and thus, provides direct comparisons with NLA, \textbf{2)}  LFADS has been frequently employed in this field, and \textbf{3)} NDT is chosen since the masked autoencoder has been proposed for a de facto standard of representation learning in many domains \cite{devlin2018bert, he2022masked, tamkin2022dabs}. 

For VAEs, the weight (\(\beta\)) for the KL-divergence term is searched over [\(10^{-4},10^{-3},10^{-2},10^{-1}\)].\footnote{Additional \(\beta=1\) is tried for LFADS.} We also train LFADS with two different sizes of the latent factor, 32 and 256. NDT is implemented with six layers of Transformer, and the representation of each layer is considered separately. The best layer is chosen by evaluation on the validation data. All the resulting representations have a size of 256 if not specified. See Appendix \ref{app:baseline} for the implementation details.

\subsubsection{NLA variants}
\label{nlavariants}
To show the effectiveness of the cross alignment loss in general, we suggest two variants of our framework. These variants share the objective but differ in the temporal alignment method, replacing the suggested TWM.

\begin{itemize}
\item \textbf{NLA-sDTW}: Instead of TWM, optimal time warping paths are found by DTW. Training with deterministic alignment from DTW is unstable, thus we use soft-DTW \cite{cuturi2017soft} to make the alignment differentiable. However, soft-DTW calculates the loss by tracking down DTW algorithm, and does not directly provide the alignment distribution nor output the warped series. Therefore, we use Equation \ref{eq:dist} for the distance function to approximate Equation \ref{eq:contrastive_loss3}. 

\begin{equation}
\label{eq:dist}
\text{sdtw\_dist}(c_i^x,s_j^{x'}) := -\text{log}\: \frac{\text{sim}(c_i^x,s_j^{x'})}{\sum_k^{L_x} \text{sim}(c_k^x,s_j^{x'})}
\end{equation}

\item \textbf{NLA-SUP}: Instead of finding alignment in unsupervised way, NLA-SUP employs a supervised alignment, which is obtained by applying DTW to AKT labels. 
\end{itemize}

\begin{figure*}[t]
\begin{center}
\centerline{\includegraphics[width=450pt]{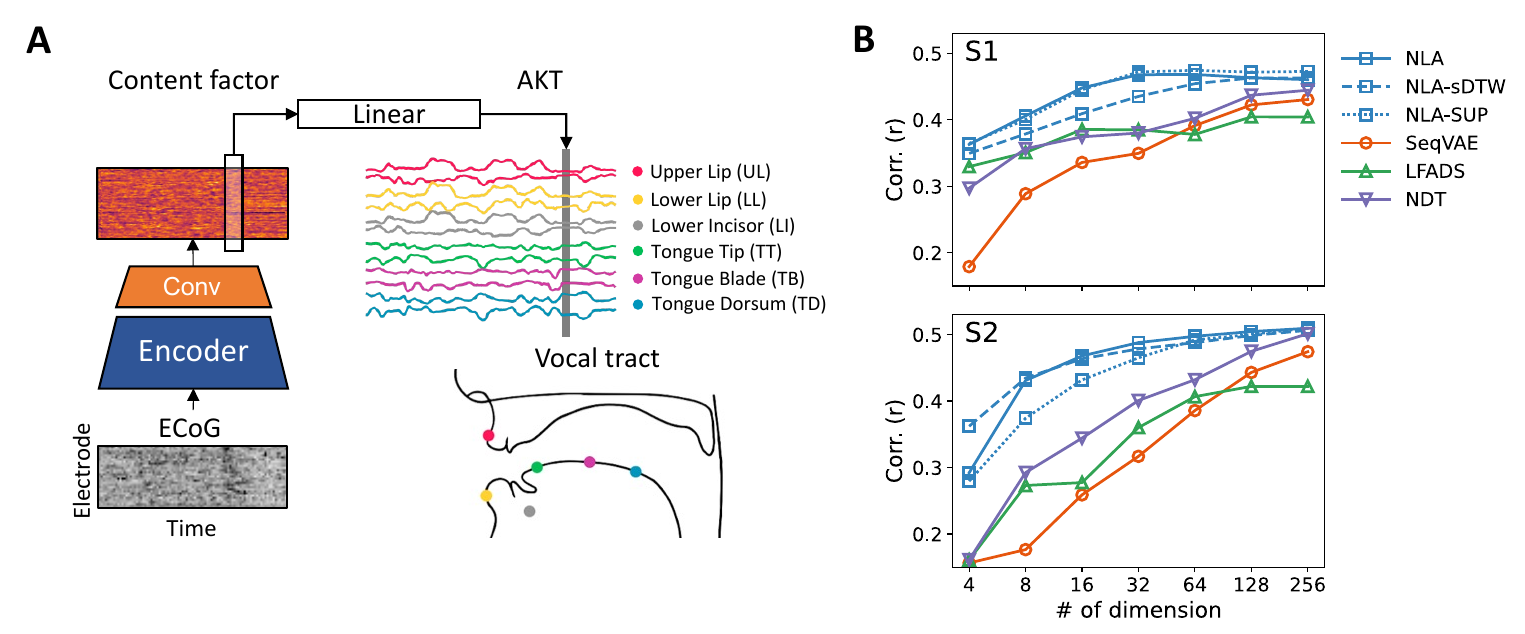}}
\caption{\textbf{A}) Overview of behavioral decoding analysis. A linear model is fitted to predict each frame of AKT, for each of XY coordinates of six articulators.  \textbf{B}) Behavioral decoding performance by reducing dimensionality (blue: NLA, blue-dashed: NLA-sDTW, blue-dotted: NLA-SUP, orange: SeqVAE, green: LFADS, and purple: NDT) The Y-axis means the decoding performance measured as averaged correlation coefficient (r), and the X-axis means the number of dimensionalities. (top: S1, bottom: S2)}
\label{behav_decoding}
\end{center}
\end{figure*}

\section{Results}


\subsection{Behavioral relevance of learned representations} \label{sec:dec}

The representations from the trained models are  evaluated by the performance in linear decoding of behavior \cite{pei2021neural}. Here, we use the articulation (AKT) as the reference behavior. AKT is the most prominently represented feature in the brain while speaking \cite{chartier2018encoding,anumanchipalli2019speech}, and linear decoding of AKT is a robust and valid task for evaluating speech representations \cite{cho2022probe}. Together, AKT is an optimal target to inspect the behavioral relevance of the neural representation of the speech. We fit ridge regression to predict AKT from 0.4-second window of content factors, and the correlation coefficients on the unseen test sentences are averaged over articulators (Figure \ref{behav_decoding} A). We also measure this score on the low-dimensional space of the learned representations, where the dimensionality is reduced by PCA.\footnote{For LFADS, the models trained with explicitly reduced dimensionality are reported if they are better than PCA. See Appendix \ref{sec:pca_reason} for further discussion.} The results are reported in Table \ref{tab:decoding}.

As shown in Table \ref{tab:decoding}, NLA and its variants show higher correlations with AKT than the baselines, for both S1 and S2. The difference from the baselines gets even larger in lower dimensional settings. When we measure AKT correlations by dimensionality, our NLAs are resistant to dimensionality reduction, while the performance of the baselines sharply declines (Figure \ref{behav_decoding} B). 

 \begin{table}[t]

\caption{Performance (r) of linear AKT decoding with full dimensionality (d=256) and reduced dimensionality (d=32).}
\vskip 0.1in
\label{tab:decoding}
\begin{center}
\begin{tabular}{l|rr|rr}
\hline

\multicolumn{1}{c|}{\multirow{2}{*}{Model}} & \multicolumn{2}{c|}{S1} & \multicolumn{2}{c}{S2}\\ \cline{2-5} \multicolumn{1}{c|}{} & \multicolumn{1}{c}{d=32} & \multicolumn{1}{c|}{d=256} & \multicolumn{1}{c}{d=32} & \multicolumn{1}{c}{d=256} \\ \hline\hline
SeqVAE & 0.349 & 0.430 & 0.318 & 0.474\\
LFADS & 0.385 & 0.404 & 0.360 & 0.422\\
NDT & 0.396 & 0.444 & 0.401 & 0.501\\ \hline
NLA  & \textbf{0.468}  & 0.460 & \textbf{0.488}  & \textbf{0.509} \\
NLA-sDTW & 0.435 & \textbf{0.463}  & 0.478 & 0.506 \\ \hline
NLA-SUP & \textbf{0.472} & \textbf{0.473} & 0.465 & \textbf{0.510}\\ \hline
\end{tabular}
\end{center}
\vspace{-4mm}
\end{table}
 
 In general, the performance variance within NLAs is lower than the difference from the baselines. Among NLAs, NLA shows the minimum loss or even gain after the dimensionality is reduced to 32. These results suggest that the cross-trial alignment learns to represent behaviors in a more compact and efficient way, and NLA can learn better content factors than NLA-sDTW, which are comparable to NLA-SUP.

\begin{figure*}[ht]
\begin{center}
\centerline{\includegraphics[width=\textwidth]{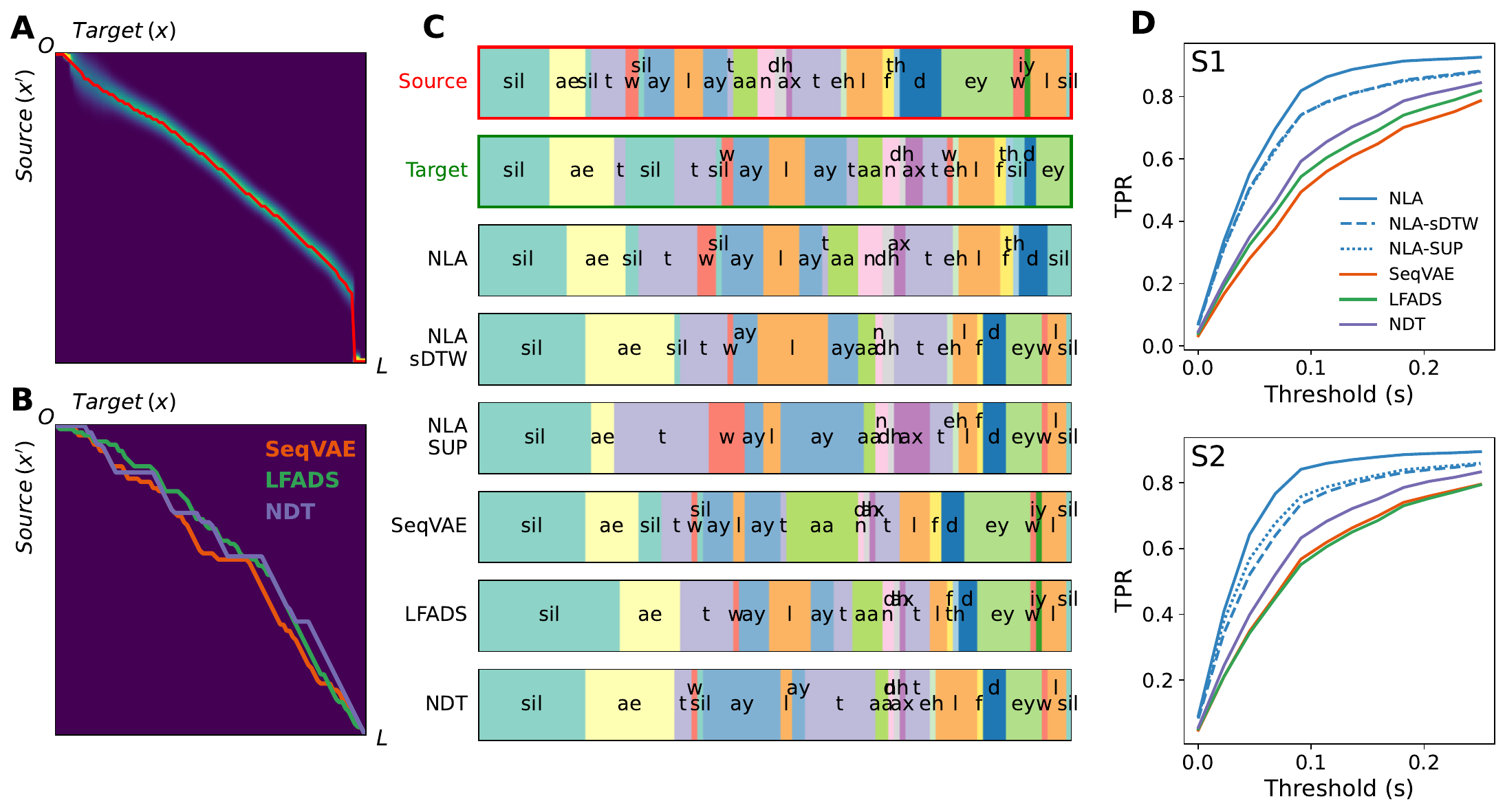}}
\caption{\textbf{A)} Example distribution of alignment inferred from TWM. The red line is the warping path estimated by taking the maximum. \textbf{B)} Alignments from DTW applied to SeqVAE, LFADS, and NDT, on the same source-target pair as A). \textbf{C)} Frame-wise phoneme labels of source (row 1), target (row 2), and source warped by the unsupervised alignment methods (row 3-8). The phoneme is annotated for each color-coded segment. The accuracy of the alignment can be qualitatively measured by inspecting vertical alignment of the segments. \textbf{D)} TPR (Y-axis) by time threshold (X-axis) for each subject (top: S1, bottom: S2). The color codes are the same as Figure \ref{behav_decoding}. }
\label{phone_alignment}
\end{center}
\end{figure*}

\subsection{Empirical validation of time warping model} \label{sec:behav_co}

As a proxy for evaluating the alignment inference, we measure the behavioral coherence between aligned trials. We create additional validation and test datasets by sampling 2-second windows of source-target pairs (valid: 200 pairs, test: 1000 pairs) for each participant. To ensure that the sampled pairs have overlapping contents, we use the supervised alignment of trials to anchor the sampling. \footnote{The same sampling method as \S \ref{online}.} Then, we apply unsupervised warping to align the source to the target and check how the phonemes and AKTs are accurately aligned.

When a phoneme in the source is warped, the distance from the reference phoneme in the target trial is measured. Then, the alignment is considered to be correct if the distance is below a threshold. As Figure \ref{phone_alignment} D, a curve of true positive rate (TPR) can be plotted by adjusting the threshold value, and the area under the curve (AUC\(:= \text{area}/\text{threshold}\)) indicates the sensitivity of the alignment. For AKT, we simply measure the average correlation between the target AKT  and the warped source AKT. We additionally report the score by the supervised alignment as a reference for the upper bound of the suggested metrics.

\begin{table}[t]
\label{tab:alnphoneme}
\caption{Aligned phoneme AUC of alignment methods.}
\vskip 0.1in
\begin{center}
\begin{tabular}{l|rr|rr}
\hline
\multirow{2}{*}{Model} & \multicolumn{2}{c|}{S1}                                  & \multicolumn{2}{c}{S2}                                  \\ \cline{2-5} 
                       & \multicolumn{1}{c}{\(\leq0.5s\)} & \multicolumn{1}{c|}{\(\leq2s\)} & \multicolumn{1}{c}{\(\leq0.5s\)} & \multicolumn{1}{c}{\(\leq2s\)} \\ \hline \hline
SeqVAE  & 0.683   & 0.874  & 0.702   & 0.868\\
LFADS  & 0.711   & 0.882   & 0.697   & 0.867\\
NDT  & 0.734   & 0.888     & 0.737   & 0.877  \\ \hline
NLA & \textbf{0.834}  & \textbf{0.927}  & \textbf{0.825} & \textbf{0.908}\\
NLA-sDTW  & 0.793     & 0.904     & 0.777     & 0.889\\ 
NLA-SUP   & 0.789 & 0.900  & 0.785   & 0.889\\   \hline
Supervised & 0.867 &  0.932  & 0.849  & 0.913 \\ \hline
\end{tabular}
\end{center}
\vspace{-4mm}
\end{table}

For our proposed TWM, the alignment is inferred by taking the centers of the output distributions (red line in Figure \ref{phone_alignment} A). For other models, the alignment is obtained by applying DTW on the representations.\footnote{See Appendix \ref{sec:behav_co_app} for selecting distance function for DTW.} The alignment inferred by TWM is smooth (Figure \ref{phone_alignment} A) but those of other models show a staircase pattern (Figure \ref{phone_alignment} B). 
When visually inspected in Figure \ref{phone_alignment} C, the source phonemes aligned by NLA (the 3rd row) show the highest coherence with the target phonemes (the 2nd row). The threshold-TPR curves of NLA are generally above those of other models, and NLA shows the highest AUC for both S1 and S2 (Table \ref{tab:alnphoneme}). The correlations of the aligned AKT are also the highest in NLA (Table \ref{tab:aktaln}). This suggests that TWM provides the most accurate unsupervised alignment of behaviors among the methods compared.

\begin{table}[t]
\caption{Aligned AKT correlation of alignment methods.}
\vskip 0.1in
\label{tab:aktaln}
\begin{center}
\begin{tabular}{l|r|r}
\hline
\multicolumn{1}{c|}{Model} & \multicolumn{1}{c|}{S1} & \multicolumn{1}{c}{S2} \\ \hline \hline
SeqVAE                     & 0.469                   & 0.495                  \\
LFADS                      & 0.483                   & 0.465                  \\
NDT                        & 0.555                   & 0.531                  \\ \hline
NLA                & \textbf{0.708}              & \textbf{0.715}                  \\
NLA-sDTW                   & 0.667                   & 0.634                  \\
NLA-SUP                    & 0.632                   & 0.657                  \\ \hline
Supervised & 0.874  & 0.851 \\\hline

\end{tabular}
\end{center}
\vspace{-4mm}
\end{table}

\begin{table}[t]
\caption{Cross-trial consistency (CTC) of learned representations.}
\vskip 0.1in
\label{tab:crosstrial}
\begin{center}
\begin{tabular}{l|r|r}
\hline
\multicolumn{1}{c|}{Model} & \multicolumn{1}{c|}{S1} & \multicolumn{1}{c}{S2} \\ \hline \hline
SeqVAE                     & 0.053                   & 0.046 (0.053)           \\
LFADS                      & 0.121 (0.216)            & 0.199 (\textbf{0.265})           \\
NDT                        & 0.091 (0.178)            & 0.182 (0.211)           \\ \hline
NLA              & \textbf{0.281}                   & \textbf{0.230}                \\
NLA-sDTW                   & 0.227                   & 0.186                  \\ \hline
NLA-SUP                    & 0.273                   & \textbf{0.242}                  \\ \hline
\end{tabular} 
\end{center}
\vspace{-4mm}
\end{table}

\begin{figure*}[ht]
\label{tsne}

\begin{center}
\centerline{\includegraphics[width=\textwidth]{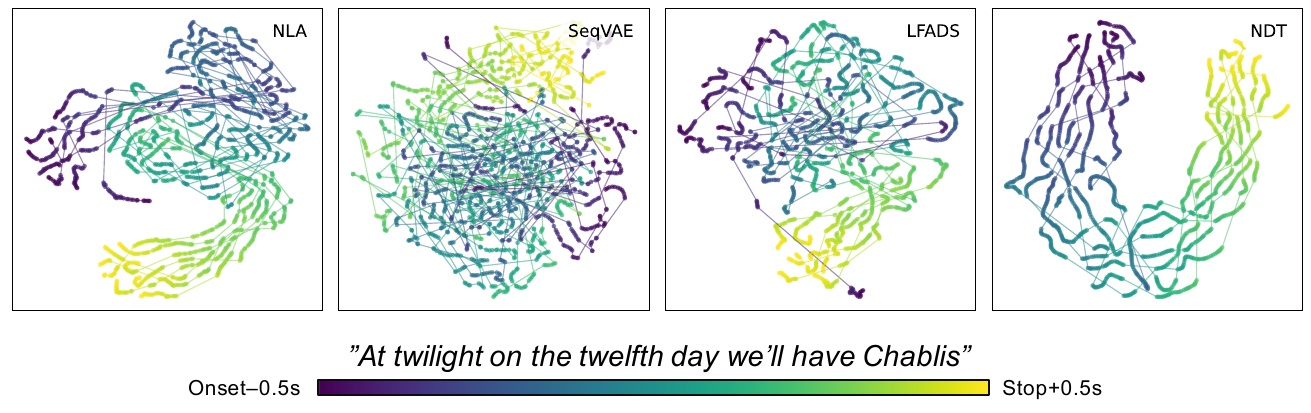}}
\caption{Manifolds visualized by t-SNE (from left to right: NLA, SeqVAE, LFADS, and NDT). The longest sentence with eight repetitions in the test set of S1 is selected: ``At twilight on the twelfth day we'll have Chablis." The progress from onset-0.5s to stop+0.5s is colored for each trial individually, and time points in the trial are connected with line. For each model class, the representation with the highest CTC is selected. }
\label{tsne}
\end{center}
\end{figure*}

\subsection{Cross-trial consistency} \label{sec:ctc}

We evaluate the cross-trial consistency of the representations to check how well they are aligned within the same behaviors. To make the metric comparable across models, we measure weighted correlations in the PCA space. The correlation of the target (\(z^A\)) and the source (\(z^B\)) is measured on each PC (e.g., the \(i\)-th PC) as \(\rho_i = \text{corr}(z^A v_i,z^B v_i)\), and then weighted by the variance of the data explained by the PC (\(w_i\)). Finally, the cross-trial consistency (CTC) is defined as summation of the weighted PC correlations (\(\text{CTC} (Z) := \sum_i^d w_i \rho_i\)). The resulting score lies in [-1, 1]. Here, we use the supervised alignment (i.e., DTW on AKT) for all models, to make the evaluation fair. The CTC for each model class is reported on Table \ref{tab:crosstrial}. Since there would be a trade-off between representation capacity and cross-trial consistency, we select hyperparameters that yield the highest decoding performance or the highest CTC. If different, we report the latter in parentheses (Table \ref{tab:crosstrial}). 

For S1, NLA shows the highest CTC, which is even higher than the supervised version (NLA-SUP). For S2, the CTC score of NLA is also close to that of NLA-SUP, and higher than the baselines except for LFADS with 0.265. This particular LFADS instance has a decoding performance of 0.327, which is significantly low based on Table \ref{behav_decoding}. Compared to NLA-sDTW, NLA in both participants outperforms NLA-sDTW by a large margin. These results demonstrate that our NLA can learn representations consistent across trials.

When the manifold is visualized by t-SNE, shared neural trajectories and dynamics are easily identifiable in NLA (Figure \ref{tsne} leftmost panel). NDT also shows a general trend (Figure \ref{tsne} rightmost panel) but with coarse dynamics compared to NLA. The representations near the stop are clustered in SeqVAE (Figure \ref{tsne} second left panel) but no visible cluster is found in other time points. LFADS show some shared dynamics but not as clear as NLA (Figure \ref{tsne} second right panel). More examples can be found in Appendix \ref{app:tsne}. 

\section{Discussion}

The content factor extracted by our model encodes behaviors with compact dimensionality. This property has the potential to enhance the sample efficiency and robustness of BCI applications, which is crucial given the expense and limited nature of data collection, especially in invasive recording. 

The definition of consistency used in our framework is flexible, so that it can be defined to be across subject to investigate neurotypical principles, or across session to stabilize long term deployment of BCI \cite{ dyer2017cryptography, karpowicz2022stabilizing}. Tasks in other domain, such as action recognition, can be revisited with our framework. Exploring these broader applications is our future direction.


Our method reveals a more interpretable and meaningful manifold than the previous methods, although the high-density ECoG is a challenging modality. Therefore, the method can be utilized to investigate the neural manifolds and dynamics of complex behaviors, without relying on external labels. As far as we know, this is the first demonstration of the unsupervised discovery of a clean manifold from highly noisy and complex neural data.

\textbf{Limitation}: Our approach requires multiple repetitions for the same condition, which may not be applicable to some experiments (e.g., spontaneous speech).

\section{Conclusion}

We propose a novel unsupervised learning framework to learn representation from noisy neural data by cross-trial alignment. To this end, we develop a contrastive alignment loss and a differentiable time warping model. The effectiveness of our proposed framework is empirically demonstrated with challenging human ECoG data. 

\section{Acknowledgements}

This research is supported by the following grants to PI Anumanchipalli --- NSF award 2106928, Rose Hills Foundation and Noyce Foundation.

\bibliography{example_paper}
\bibliographystyle{icml2023}

\newpage
\appendix
\onecolumn
\section{Implementation detail}
\label{app:A}
\subsection{Sequential autoencoder}
\label{app:cnn}

The backbone of our proposed famework is a sequential autoencoder with 1D CNN. Both encoder and decoder are composed of two layers of convolutional layers with residual connection (ResLayer). The configurations are shown in the Table \ref{app:archi}. The parameters are denoted as the PyTorch convention: conv1d (in channel,out channel, kernel, stride, dilation), and convT1d (in channel, out channel, kernel, stride). The mapping, f, is defined as two stacks of conv1d (256, 256, 3, 1, 1) on top of the encoder. The activation function, ReLU, is applied for the output of every conv1d, and followed by Batch normalization. The input to this model is in the shape of (Batch size (B), T=400, number of channels=256), and the resulting latent factors have (B, L=100, d=256).

\begin{table}[!h]
\label{app:archi}
\caption{Architecture of the backbone sequential autoencoder.}
\begin{center}
\begin{tabular}{cc}
\hline
\multicolumn{2}{c}{Encoder}                                                  \\ \hline \hline
\multicolumn{1}{c|}{\multirow{2}{*}{ResLayer1}} & conv1d (256, 256, 3, 2, 1) \\ \cline{2-2} 
\multicolumn{1}{c|}{}                           & conv1d (256, 256, 3, 1, 2) \\ \hline
\multicolumn{1}{c|}{\multirow{2}{*}{ResLayer2}} & conv1d (256, 256, 3, 2, 1) \\ \cline{2-2} 
\multicolumn{1}{c|}{}                           & conv1d (256, 256, 3, 1, 2) \\ \hline
\multicolumn{1}{c|}{Linear}                     & fc (256, 256)               \\ \hline
\multicolumn{2}{c}{Decoder}                                                  \\ \hline \hline
\multicolumn{1}{c|}{Upsample}                   & convT1d (256, 256, 2, 1)       \\ \hline
\multicolumn{1}{c|}{\multirow{2}{*}{ResLayer1}} & conv1d (256, 256, 3, 1, 1) \\ \cline{2-2} 
\multicolumn{1}{c|}{}                           & conv1d (256, 256, 3, 1, 1) \\ \hline
\multicolumn{1}{c|}{Upsample}                   & convT1d (256, 256, 2, 1)       \\ \hline
\multicolumn{1}{c|}{\multirow{2}{*}{ResLayer2}} & conv1d (256, 256, 3, 1, 1) \\ \cline{2-2} 
\multicolumn{1}{c|}{}                           & conv1d (256, 256, 3, 1, 1) \\ \hline
\multicolumn{1}{c|}{Logit}                      & fc (256, 256)               \\ \hline
\end{tabular}
\end{center}
\end{table}

\subsection{Time warping model}
\label{app:transformer}
The time warping model is implemented with four layers of Transformer \cite{vaswani2017attention}. We follow a conventional practice of implementing Transformer. Each Transformer layer is composed of multi-head attention and feed forward modules. Following the same notation in \citet{vaswani2017attention}, we use \(d_{\text{model}}=64, d_k=32, h=4\) for multi-head attention, and \(d_{\text{ff}}=128\) for the feed forward model. ReLU is used for the activation function of the feedforward model, and Dropout rate 0.2 is applied to the attentions and the output of the first layer in the feedforward model. Additional Layer normalization is applied to the output of each module. The proposed time warping model is then composed of an input layer with fc(256, 64), four stacks of the Transformer blocks, and an output layer with fc(64, 2). The positional embedding (PE) and sequential embedding (SE) are added to the input before the Transformer layers. The input to this model is pairs of sequences, [(B, L, d), (B, L, d)], and then the output is also pairs of parameters for alignment distribution [(B, L, 2), (B, L, 2)]. In the main text, we suggest to warp one trial to another using one half of the outputs, but in practice, we align them bidirectionally by leveraging the other half part of the outputs. 

For soft-DTW in NLA-sDTW, we use codes provided by \citet{maghoumi2021deepnag}, using default setting (\(\gamma=1.0\)) other than the proposed distance function. For inferring alignment after training the models except for NLA, we use DTW implemented by \citet{giorgino2009computing} and every DTW uses default setting without window.

\subsection{Training configuration}
\label{app:training}
We use one NVIDIA RTX A5000 to train the models, and the batch size of 64 is used. We augment the data by randomly dropping 5-10\% of channels of randomly chosen 50\% of trials in the batch. Adam optimizer is used with \(\beta_1 =0.9, \beta_2=0.999\). The number of iterations is 500K and the learning rate is annealed from 1e-3 to 1e-4 using cosine annealing. 

\section{Data preprocessing detail}
\label{app:data}

\subsection{Signal processing}
The ECoG with 16 x 16 grids collected raw local field potentials in 3K Hz. The 60 Hz line noise by DC connector is filtered out with a notch filter. Then, the Hilbert transform is applied to extract the analytic amplitude of the high-gamma frequency (70–200 Hz). Lastly, the signals are downsampled to 200 Hz. 

\subsection{Articulatory kinematic trajectory (AKT)}

The audio-to-articulation inversion (AAI) model is trained with bidirectional LSTM on two electromagnetic articulography (EMA) datasets,  MOCHA-TIMIT \cite{mochatimit} and MNGU0 \cite{richmond2011mngu}. The AAI model is trained to predict 12-dimensional (X,Y coordinates for each of 6 articulators) articulatory representations (AKTs) from acoustic features (MFCC). Then, the AKTs are inferred using recorded participants' audio. 

\subsection{ECoG grid locations}

\begin{figure*}[ht]
\begin{center}
\centerline{\includegraphics[width=300pt]{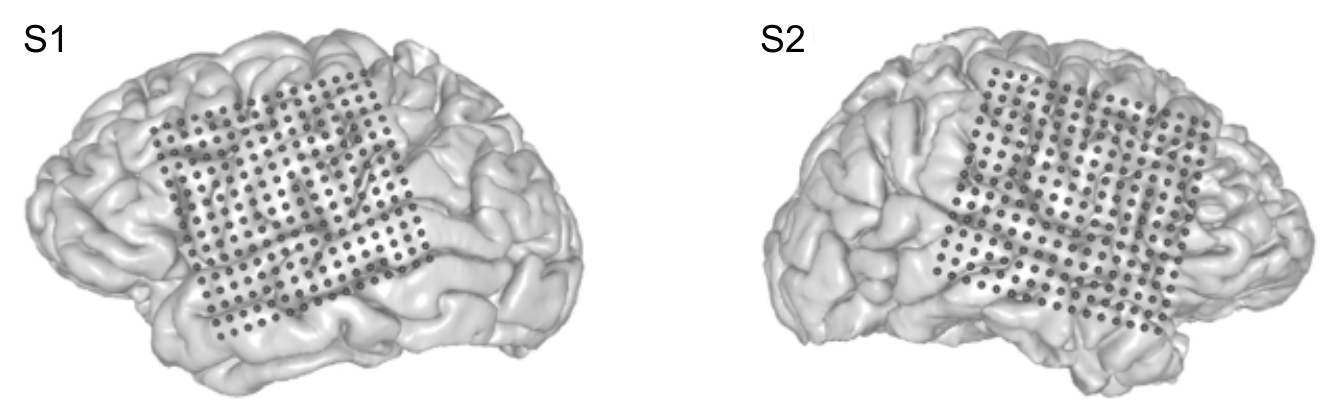}}
\caption{The location of the implanted ECoG (left: S1, right: S2). The figures are from \citet{anumanchipalli2019speech}}
\end{center}
\end{figure*}

\section{Baseline implementation}
\label{app:baseline}
\subsection{SeqVAE}
The variational inference is added to the autoencoder with the same CNN architecture as \ref{app:cnn}. A dynamic prior on the latent space is defined in non-parametric way. We follow the ideas and the implementation by the previous deep sequential VAE models \cite{li2018disentangled, zhu2020s3vae, lian2022robust}. 

For each time point, a sample from prior distribution is generated by running LSTM on random vectors sampled from a diagonal Gaussian distribution, \(\mathcal{N}(0, I)\). This LSTM is trained along with the VAE and provides sample depend on the sampling history, shaping dynamical prior of the embedding space. We redirect to \cite{li2018disentangled} for the detailed formulation of the dynamic prior. This LSTM is implemented as a single layer with the hidden size of 256.

We train model for different weighting factors of KL-divergence term in ELBO:  [1e-4, 1e-3, 1e-2, 1e-1]. All the training configurations are the same as \ref{app:training}. The batch size is increased to 128 since the pairs are batched in NLA, so we need to double up the batch size to match the size. 

\subsection{LFADS}

We follow the implementation by \citet{pandarinath2018inferring, keshtkaran2022large}, with a bidirectional GRU encoder, and a unidirectional GRU controller and 
generator. The size of all the hidden states are set as 128, and the size of the factor is set as 32 or 256. For the prior on the initial state, we used a Gaussian distribution with trainable mean and fixed variance of 0.1. The prior on the controller output is autoregressively defined with trainable autocorrelation variables. The input signals are downsampled to 50 Hz to match frequency of the resulting representation from other models. The weight for KL-divergence term is selected from  [1e-4, 1e-3, 1e-2, 1e-1, 1], and fixed throughout the training. The number of iterations is set as 50K and the batch size is also doubled. Other settings are identical to \ref{app:training}.

\subsection{NDT}

For NDT, a masked autoencoder with the Transformer architecture is implemented. There are some key differences from the original model proposed by \citet{ye2021representation}: 1) convolutional positional encoding, 2) randomized masking plan, and 3) modified loss objective to incorporate \emph{unmasked} prediction (autoencoding). 

First of all, since Transformer lacks generalizability to unseen lengths, the model trained with the fixed window of inputs fails to model variable length of trials. To tackle this, our NDT takes the full-length trials as inputs, and to do so, we use convolutional positional embedding \cite{baevski2020wav2vec} instead of the lookup table positional embedding. The convolution with kernel size of 128 and group of 8 is applied to the input to extract positional information from the data, and then added to the input before entering the Transformer layers. This convolutional position encoding is proven to be successful in speech domain \cite{baevski2020wav2vec} and agnostic to the input length. Other than that, Transformer has the same architecture as \ref{app:transformer} with \(d_{\text{model}}=256,h=4, d_k=64, d_{\text{ff}}=256\), and six layers are used.

We also diversify the masking probability by randomly selecting from [0.2, 0.4, 0.6]. The input sequence are first segmented and each segment is masked with the chosen masking probability. The length of the segment is randomly chosen from [5, 10]. The model overfits severely if only trained with masked prediction objective. Therefore, we use an auxiliary loss of the reconstruction of the unmasked part. This loss is weighted with 0.1 and added to the masked prediction loss. This modification prevents the model from overfitting. The number of iterations are set as 100K and the other settings are the same as \ref{app:training}.

\section{Additional analyses}

\subsection{Reconstruction performance}

The ECoG signals are highly noisy and there is spontaneous background activity that is not random but behaviorally irrelevant. Thus, 
 we don't compare this in the main body since the reconstruction performance doesn't distinguish noise, failing to be a valid evaluation metric. Table \ref{tab:recon} shows the reconstruction performance of each model, measured as average correlation coefficients of 256 channels. We select the model instance by decoding accuracy in validation set and the number in parentheses means the score selected using decoding from low-dimensional models (dim=32) if applicable. The reconstruction performance is maxing out in NLAs and this is natural since the content factor is not directly involved in the reconstructions. So NLA is free of the capacity-regularization trade-off that is shown in the case of SeqVAE and LFADS.

\begin{table}[ht]
\caption{Reconstruction accuracy (r)}
\vskip 0.1in
\label{tab:recon}
\begin{center}
\begin{tabular}{l|r|r}
\hline
\multicolumn{1}{c|}{Model} & \multicolumn{1}{c|}{S1} & \multicolumn{1}{c}{S2} \\ \hline \hline
SeqVAE                     & 0.959 (0.902)                & 0.963 (0.909)                  \\
LFADS                      & 0.721 (0.528)                & 0.549 (0.765)                  \\
NDT                        & 0.907                       & 0.918                  \\ \hline
NLA                        & 0.992                       & 0.985                  \\
NLA-sDTW                   & 0.988                        & 0.984                  \\\hline
NLA-SUP                    & 0.990                       & 0.986                  \\ \hline
\end{tabular}
\end{center}
\vspace{-4mm}
\end{table}

\subsection{Manifold visualization}
\label{app:tsne}

\begin{figure*}[ht]
\label{tsne_all}
\begin{center}
\centerline{\includegraphics[width=440pt]{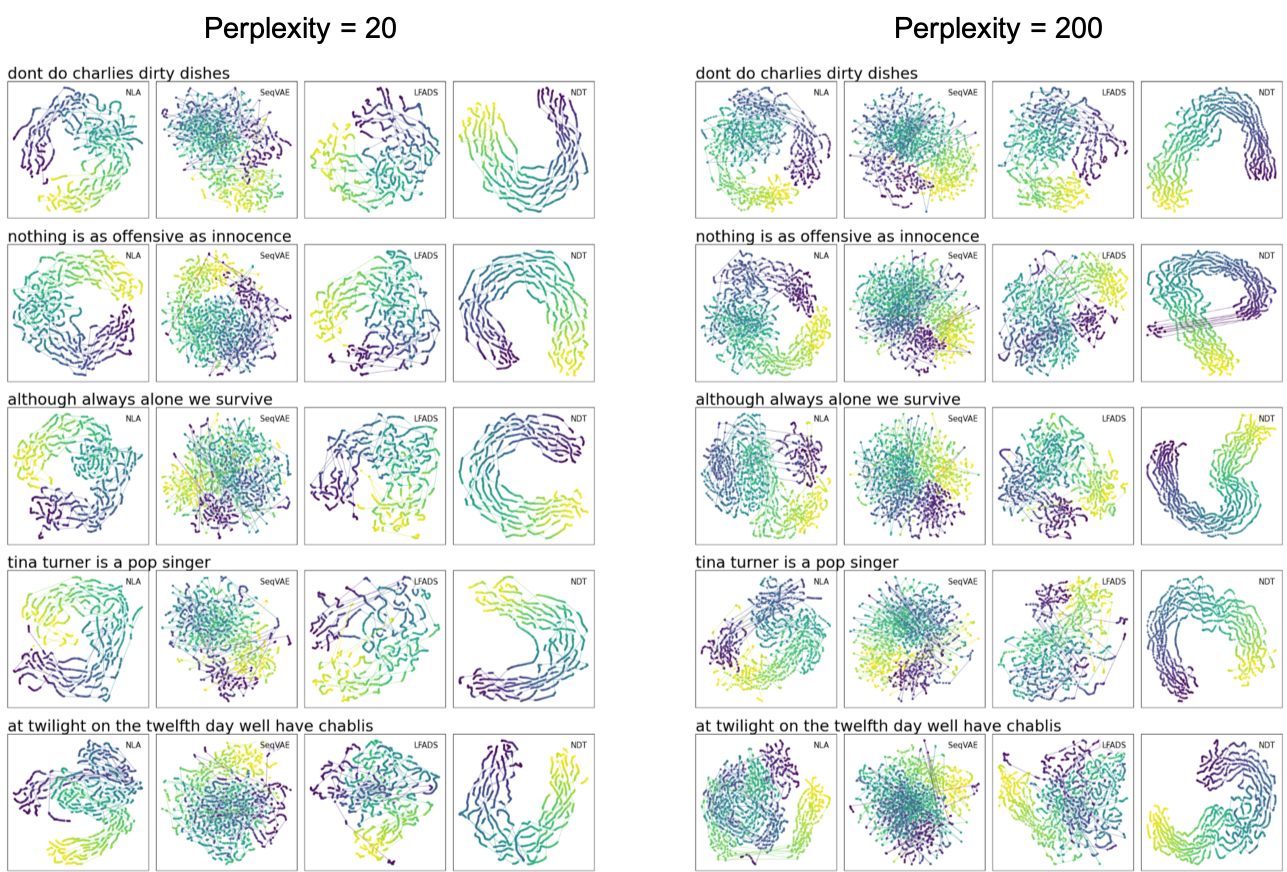}}
\caption{t-SNE visualizations on other test sentences with different perplexities (left: 20, right: 200). The main figure uses perplexity=20.}
\end{center}
\end{figure*}

The manifolds of other test sentences in S1 are visualized in Figure \ref{tsne_all}. The difference across models pointed out in the main text is replicated in other sentences. We also try with higher perplexity (right) and the results show the same conclusion. The dimension of the data is reduced to 32 by PCA prior to running t-SNE.

\section{Other design choices}

\subsection{Distance functions for training NLA} \label{sec:dist_exp}

The choice of distance function affects the stability of training the TWM in NLA. The instability of TWM results in degeneration of the inferred alignment (e.g., all the alignment center indices are collapsed to zero or the end). With the Euclidean distance or cosine distance, the training requires more careful hyperparameter selection. However, training TWM with the inner product distance was highly stable. Thus, we decided to use the inner product as the similarity function for all experiments. For further stabilization, the inner product values are softly clamped to be in [-5, 5] by \(5\: \text{tanh}(x/5)\). Instead, we experimented with the Euclidean distance and cosine distance for one of our variants, NLA-sDTW. For cosine distance, the similarity is divided by temperature which is set as 0.2. We consistently found lower performance when cosine or the Euclidean distance is used. Table \ref{tab:dist_fun} shows the scores of NLA-sDTW with other distance functions: decoding (\S \ref{sec:dec}), alnPh \& alnAKT (\S \ref{sec:behav_co}), and CTC (\S \ref{sec:ctc}).

\begin{table}[ht]
\caption{Model performance by distance function type}
\vskip 0.1in
\label{tab:dist_fun}
\centering
\begin{tabular}{l|ll|ll|ll|ll} 
\hline
\multirow{2}{*}{Distance} & \multicolumn{2}{c|}{decoding}                                                                         & \multicolumn{2}{c|}{alnPh}                                                                            & \multicolumn{2}{c|}{alnAKT}                                                                           & \multicolumn{2}{c}{CTC}                                                                                \\ 
\cline{2-9}
                          & \multicolumn{1}{c}{S1}                                                & \multicolumn{1}{c|}{S2}                                                &\multicolumn{1}{c}{S1}                                                & \multicolumn{1}{c|}{S2}                                                & \multicolumn{1}{c}{S1}                                                & \multicolumn{1}{c|}{S2}                                                & \multicolumn{1}{c}{S1}                                                & \multicolumn{1}{c}{S2}                                                 \\ 
\hline\hline
product                   & \textcolor[rgb]{0.173,0.227,0.29}{\textbf{0.463}} & \textcolor[rgb]{0.173,0.227,0.29}{\textbf{0.506}} & \textcolor[rgb]{0.173,0.227,0.29}{\textbf{0.904}} & \textcolor[rgb]{0.173,0.227,0.29}{\textbf{0.889}} & \textcolor[rgb]{0.173,0.227,0.29}{\textbf{0.667}} & \textcolor[rgb]{0.173,0.227,0.29}{\textbf{0.634}} & \textcolor[rgb]{0.173,0.227,0.29}{\textbf{0.227}} & \textcolor[rgb]{0.173,0.227,0.29}{\textbf{0.186}}  \\ 
\hline
cosine                    & 0.457                                             & 0.498                                             & 0.895                                             & 0.883                                             & 0.617                                             & 0.596                                             & 0.083                                             & 0.098                                              \\ 
\hline
euclidean                 & 0.442                                             & 0.503                                             & 0.883                                             & 0.865                                             & 0.596                                             & 0.453                                             & 0.034                                             & 0.042                                              \\
\hline
\end{tabular}
\end{table}

\subsection{Optimal distance function for post-training DTW} \label{sec:behav_co_app}

As the temporal alignments are not inferred in the baseline models, we use DTW to get the temporal alignments between trials after training the models. We experimented with the Euclidean distance and product distance (negative of inner product) particularly for this post-training DTW. Table \ref{tab:post_dtw} shows behavioral coherence scores, alnPh and alnAKT (\S \ref{sec:behav_co}), by distance functions for each model. For the baseline models (SeqVae, LFADS, NDT), the product distance is mostly selected as optimal on the validation set. For NLA-sDTW and NLA-SUP, the Euclidean distance is mostly selected as optimal. However, the difference in the scores across distance methods is not as significant as the difference between the scores of NLAs and those of the baselines. The choice of distance has minimal effect in post-training DTW and the alignments inferred by TWM show higher scores than any configurations.

\begin{table}[ht!]
\caption{Behavioral coherence by distance function type in post-training DTW.}
\vskip 0.1in
\label{tab:post_dtw}
\centering
\begin{tabular}{l|l|ll|ll} 
\hline
\multirow{2}{*}{Model }   & \multirow{2}{*}{Distance} & \multicolumn{2}{c|}{alnPh}      & \multicolumn{2}{c}{alnAKT}       \\ 
\cline{3-6}
                          &                           & \multicolumn{1}{c}{S1}               & \multicolumn{1}{c|}{S2}               & \multicolumn{1}{c}{S1}               & \multicolumn{1}{c}{S2}                \\ 
\hline\hline
\multirow{2}{*}{SeqVAE}   & product                   & \textbf{0.874} & \textbf{0.868} & \textbf{0.469} & \textbf{0.495}  \\ 
\cline{2-6}
                          & euclidean                 & 0.868          & 0.862          & 0.444          & 0.449           \\ 
\hline
\multirow{2}{*}{LFADS}    & product                   & \textbf{0.882} & \textbf{0.867} & 0.483          & \textbf{0.465}  \\ 
\cline{2-6}
                          & euclidean                 & 0.871          & 0.859          & \textbf{0.486} & 0.460           \\ 
\hline
\multirow{2}{*}{NDT}      & product                   & 0.887          & \textbf{0.877} & \textbf{0.555} & \textbf{0.531}  \\ 
\cline{2-6}
                          & euclidean                 & \textbf{0.888} & 0.873          & 0.531          & 0.521           \\ 
\hline
\multirow{2}{*}{NLA-sDTW} & product                   & 0.897          & 0.889          & 0.647          & 0.634           \\ 
\cline{2-6}
                          & euclidean                 & \textbf{0.904} & \textbf{0.891} & \textbf{0.667} & \textbf{0.639}  \\ 
\hline
\multirow{2}{*}{NLA-SUP}  & product                   & 0.895          & 0.887          & 0.632          & 0.649           \\ 
\cline{2-6}
                          & euclidean                 & \textbf{0.900} & \textbf{0.889} & \textbf{0.664} & \textbf{0.657}  \\ 
\hline
NLA                       & N/A                       & \textbf{0.927} & \textbf{0.908} & \textbf{0.708} & \textbf{0.715}  \\
\hline
\end{tabular}
\end{table}

\subsection{Reason for using PCA to reduce the dimensionality of the learned reperesentations} \label{sec:pca_reason}
We reduce dimensionality post hoc since models with explicitly reduced dimensionality tend to overfit. The overfitting problem is particularly happening with the reconstruction by the decoder since we only reduced the size of target representations, d, and left other parts the same. Thus, the decoder has the same capacity while the input has lower dimensionality. Therefore, we suspect that overfitting is happening because the decoder is trained to compensate for the neural signals from short input encoding, which is hard to be generalized. This may be mitigated by controlling the size of the decoder as well, but this opens up too much degree of freedom to cover. Thus, we decided to use post hoc dimensionality reduction by PCA due to the limitation of computing resources to explore other possibilities.


\end{document}